# HARDWARE/SOFTWARE CO-DESIGN FOR SPIKE BASED RECOGNITION

*Arfan Ghani, Martin McGinnity, Liam Maguire, Jim Harkin*

Intelligent Systems Research Centre
University of Ulster
Derry, BT487JL, N. Ireland, UK
Email: Ghani-a1@ulster.ac.uk

## ABSTRACT

The practical applications based on recurrent spiking neurons are limited due to their non-trivial learning algorithms. The temporal nature of spiking neurons is more favorable for hardware implementation where signals can be represented in binary form and communication can be done through the use of spikes. This work investigates the potential of recurrent spiking neurons implementations on reconfigurable platforms and their applicability in temporal based applications. A theoretical framework of 'reservoir computing' is investigated for hardware/software implementation. In this framework, only readout neurons are trained which overcomes the burden of training at the network level. These recurrent neural networks are termed as microcircuits which are viewed as basic computational units in cortical computation. This paper investigates the potential of recurrent neural reservoirs and presents a novel hardware/software strategy for their implementation on FPGAs. The design is implemented and the functionality is tested in the context of speech recognition application.

## 1. INTRODUCTION

This paper presents an alternative solution for parallel implementation of neural reservoirs on reconfigurable platforms (FPGAs). Software implementations of small scale neural reservoirs may not be a serious bottleneck, however to exploit the inherent parallelism of neural networks, hardware implementations are crucial. The temporal nature of bio plausible neural systems is more favorable for hardware implementation where signals can be represented in binary form and communication can be done through the use of spikes. A novel hardware/software strategy is proposed for designing neuro inspired models and their functionality is tested in the context of speech recognition application. The proposed architecture fully exploits the scalability and reconfigurability of FPGAs.

The size of network that can be implemented on a reconfigurable platform is restricted by the logic and mathematical operators available on a single device. Therefore, a hardware/software codesign strategy has to be devised for implementation of neuro inspired complex systems on reconfigurable platforms. One of the bottlenecks in implementing large scale of artificial neurons on reconfigurable platforms is the limited number of embedded multipliers available on a single device. The number of multipliers grows as the square of the number of neurons. A fully connected 2-layer network of size 10 neurons will require 100 multipliers and if the network size is increased to 100 neurons, it will require 10,000 multipliers. An area efficient architecture is presented which overcomes the burden of multipliers required for synaptic multiplications. The novel architecture was implemented as neural reservoir and overall system was tested with a speech recognition application.

## 2. RECURRENT NEURAL RESERVOIR

Networks of spiking neurons that use temporal coding are becoming more important in tasks such as computer vision, speech recognition and motor control. The results show that computation and coding in these networks is quite different and more powerful than classical networks [6].

In feedforward networks, the inputs are propagated in feed forward manner from input to the output layer. These networks are easy to analyse and standard training algorithms such as backpropagation can successfully be applied. These networks have successfully been used for static input data classification but the drawback of such networks is their inefficiency to deal with temporal data. In order to deal with temporal data, one of the possible solutions is to use recurrent networks where feedback loops are incorporated in order to flow back information for future time steps. Recurrent neural networks are promising for temporal problems but it is hard to train, control and analyse such networks. In order to overcome the burden of training in recurrent neural networks, a technique is proposed by [5],[7],[8] and termed as reservoir computing [9].

Reservoir computing is a time-efficient way to train a system for recognition related problems. It overcomes the burden of training in recurrent neural networks and facilitates it by only training the readout neurons instead of the whole recurrent network. In this paradigm, the local and global connections are constructed in a random fashion and the reservoir is perturbed with an input stream. It is important to analyse the stable reservoir dynamics by adjusting the parameters such as size, node types, input connectivity and recurrent connections. These parameters are important to examine the *short term* memory and *separation* capability of a reservoir. Once the reservoir is simulated it is left untouched and only different states are recorded for post-processing with 'readout' neurons. Training 'readout' neurons are far more efficient than training the whole recurrent network. The time required to simulate a reservoir depends on its size and the node type (neuron model) used to construct a reservoir. Simulating small networks on sequential machines may not be critical but for large networks it becomes a serious bottleneck. The simulation time increases many folds if bigger reservoirs to be simulated.

## 2.1. Neuron model

In a network of spiking neurons, each input neuron receives signals from other neurons with different synaptic strengths at different times. A single neuron is further connected with other neurons in the network through synaptic clusters. These clusters are shared amongst neurons in a network and the membrane dynamics and their corresponding spike firing times are effected by the synaptic efficacy of these clusters (see fig.1).

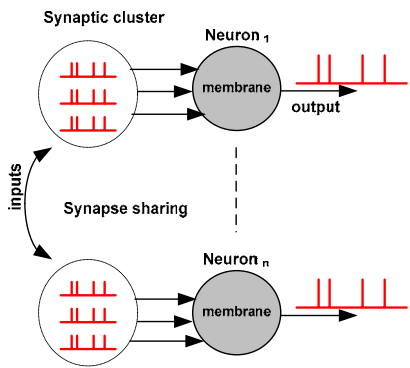

**Fig. 1**. This figure shows a synaptic interaction amongst different neurons, each neuron shares synaptic clusters from other neurons

A LIF (Leaky Integrate-and-Fire) neuron model was chosen for the implementation of neural reservoir and the mechanism of synaptic integration was modelled by the following equation:

$$V_s(t) = \sum_{i=1}^{N_s} w_i x_i(t) \qquad (1)$$

Where $V_s$ is the sum of incoming synaptic potentials to the membrane, $N_s$ is the number of synapses, $w_i$ is the synaptic efficacy and $x_i$ are the incoming binary spikes 0 or 1 at time $t$. These synaptic potentials were accumulated in a membrane and when the total synaptic potential exceeded a certain threshold, an output spike was generated.

The output of a neuron $i$ at time $t$ is modelled with the following equation:

$$O_i(t) = \begin{cases} 1, Vi(t) \geq V_{th} \\ 0, Vi(t) < V_{th} \end{cases} \qquad (2)$$

In equation 2, $V_{th}$ denotes the firing threshold and $V_i$ denotes the membrane potential of neuron $i$ at time $t$. The neuron firing dynamics were modelled with the following equation:

$$V_m(t) = \begin{cases} V_{reset} \\ V_m(t-1) + V_S(t) - V_{leakage}(t) \end{cases} \qquad (3)$$

In equation 3, $V_m$ is the membrane potential, $V_{reset}$ is the reset potential, $V_m (t-1)$ is the membrane potential at previous time step, $V_s$ is the sum of synaptic potential and $V_{leakage}$ is the exponentially decreasing leakage voltage with time constant $\tau$. The membrane voltage $V_m(t)$ will be at $V_{reset}$ if $V_s(t) > V_{th}$, otherwise the membrane potential will be equivalent to the second term in equation 3.

## 3. RECONFIGURABLE ARCHITECTURE

Hardware implementation of spike based neurons is advantageous because these neurons communicate through short pulses or so called spikes. It is generally believed that the information is conveyed through exact timing of these pulses and that the shape of the spike has no relevance to the information [10]. For an area efficient hardware implementation of biologically plausible neurons, it is necessary that the use of area hungry operators such as multipliers are minimised or completely avoided [3]. In the traditional modelling of synapses, inputs are multiplied with fixed weights and due to this multiplication the number of multipliers increases with an increase number of synapses which is a serious bottleneck for an efficient implementation of a medium to large scale networks on a single device. In the proposed design, special emphasis is given to the optimisation of the number of embedded multipliers required for the implementation of synapses.

The architecture was split into two sub architectures: synapse and membrane. For synapse modelling, a strategy is proposed where inputs were encoded in spike trains and spike counters were used to model synaptic strengths. The incoming spikes were counted and weighted through a fixed weight value. An output value of '1' is generated through a simple logic AND function when both inputs were high. As shown in fig.2, each neuron has multiple synapses where input pulses were counted and weighted through a fixed weight value. The synapse function implemented in the proposed architecture is a simple logic function of two inputs (incoming spike trains and fixed weight values). The fixed weight values were stored in the registers and random values were generated through linear feedback shift register (LFSR). The fixed weight values were compared with the randomly generated values and if the generated value equals to the fixed weight value and the number of incoming pulses were equal to the value of pulse counter then an output spike is generated, if not, no pulse will be generated. The pulse and no pulse will correspond to the inhibitory and excitatory synapse of the neuron and will contribute in the membrane potential. If an output pulse is generated, it is considered as an excitatory otherwise an inhibitory synapse. This procedure is repeated during the course of the full presentation of the input spike trains. The random weight generator block (LFSR) generates new weights at each time step in the range of plus minus 0.4. The random weight generator was modelled as a Fibonacci pseudo-random number generator with an XOR gate at the beginning of the register chain that XORs the outputs from some of the registers going into the first register. Total 6 bits were used for their implementation. Due to this simplification, the multipliers were completely avoided and synapse multiplication was modelled with a logic function of two variables *W (fixed weights)* and *I (incoming spikes)*. The weights were represented with a fixed point representation of 4 bits in a Fix_4_3 format.

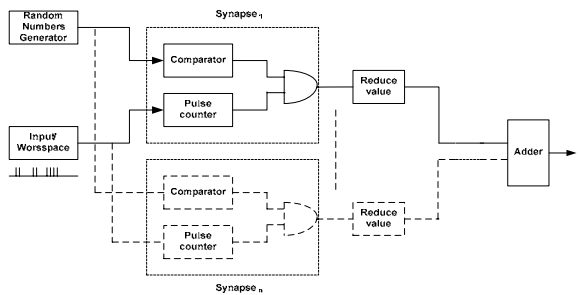

**Fig. 2**. This figure shows the modelling of synapses through pulse counting and AND gate.

The synaptic values of '1' were scaled down through shift right operations. This scaling is important for practical reasons so that enough time is given to the membrane potential to accumulate synaptic inputs and once the total membrane potential exceeded a threshold value, an output spike was generated and connected with other neurons in the network. The threshold voltage was set to 0.15 V and reset voltage to 1 mV. In the absence of spikes, the membrane potential decays exponentially to the reset voltage based on the programmable value of the decay constant. The decay constant value of -0.11 was used in these simulations.

The second half of the architecture is implemented as a neural membrane (see fig.3) where synaptic currents were accumulated in the membrane and an output spike is generated when the total membrane potential exceeded a programmable threshold $V_{th}$. The membrane of the neuron is implemented as an 18 bit accumulator with 12 bit binary points. A programmable threshold value of 0.15 V is used and after spike generation the accumulator was reset to the value of 1 mV. In the absence of input spikes, the membrane potential decays exponentially to the reset voltage and starts integrating after arrival of new incoming spikes. The exponential decay depends on the value of the programmable decay constant $\tau$. Once an output spike is fired, the neuron immediately resets to the voltage level 0 and this period is considered as refractory period during which no output spike is generated.

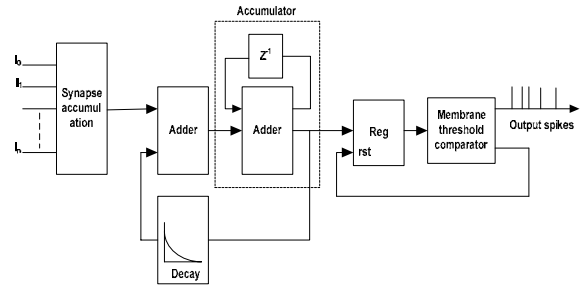

**Fig. 3**. This figure shows the modelling of a membrane where incoming synapses are accumulated and the total strength is compared against the membrane threshold.

The architecture was implemented with the Xilinx System Generator toolbox and a discrete time step of 0.125 ms is chosen for these simulations. The maximum clock speed is defined implicitly which depends on the propagation delay of the components used in the design. In Xilinx System Generator, the computational blocks receive inputs and produce outputs at every clock cycle. The Xilinx blocks were assigned computation latencies in order to match paths which have to be simulated in parallel. The fixed point simulations were used where total numbers of bits were defined along with the binary points. The fixed point format provides a flexibility to use the number of bits to represent a number. It is not area efficient to use the same fixed point representation for all the blocks in the design, therefore a format has to be chosen which is good enough to provide required precision and accuracy. For this implementation, both synapses and neuron (membrane) were represented with 18 bits in the Fix_18_12 bit format.

Other blocks such as LFSR, comparators, constant values and register delays were represented with different precision formats in order to save area. A trade-off has to be made in precision and area where higher precision will cost more area and less precision could cause errors. Once a correct functionality is achieved through simulations, the VHDL code was generated and synthesised through Xilinx ISE toolset for FPGA implementation and different hardware resources and maximum frequency was calculated.

The design was targeted for Virtex-II Pro device (xc2vp50) with a speed grade of 5. A single neuron with two synapses took 85 slices out of 23616. The design could run with a maximum clock speed of 74 MHz. The synapses were modelled without multipliers; however, one embedded multiplier will be required to model exponential decay of a leaky membrane. Total 680 slices and 8 multipliers will be required for a reservoir of size 8 neurons with 16 synapses. It takes only 8 slices to implement two synapses and if the total number of synapses were increased to 100, it will take 400 slices. If the synapses were modelled with traditional multiplication technique, then a reservoir of size 8 neurons with 16 synapses will require 24 embedded multipliers (16 for synapses and one for leaky membrane for each neuron) and by increasing the number of synapses the requirement for multipliers will increase linearly and the maximum number of synapses will be limited by the maximum number of embedded multipliers. The proposed design completely avoids the multipliers for synapses and regardless of the number of synapses only one multiplier will be required per neuron. It is possible to optimise the speed of a network with either increasing the frequency of the clock or increasing the step size. The maximum frequency allowed in a design is restricted by the maximum delay in a combinational path which is also termed as the worst case delay. The overall speed can be improved by breaking some of the longest combinatorial paths and introducing some registers. The overall speed can also be improved by increasing the step size, however care must be taken to analyse the details of the design so that spike activity is not missed during the intervals of time steps.

The resources available on a single device are not a limitation to implement parallel reservoirs. The proposed architecture offers an alternative solution for implementing one big reservoir on a single device or several compact fully parallel reservoirs. It was analysed that small reservoirs are more predictable and stable and hence can be used for parallel implementations. The fixed point format provides a flexibility to use the number of bits to represent a number and is more favorable for area efficient implementation.

## 4. PROOF OF CONCEPT

The proposed architecture was tested with an application of speech recognition where the TI46 dataset for isolated digit recognition was used which consisted of 10 isolated spoken digits from the speech corpus [11]. In order to validate the functionality of the neural reservoir, an integrated hardware/software system is proposed where signal pre and post-processing was done in software and reservoir was implemented on hardware (see fig.4). The input speech signals were pre processed to remove silence parts and features were extracted with the technique of Linear Predictive Coding. This pre-processing is very important to compress the data and only useful features were processed. These features were further converted into Poisson spike trains to be processed as an input stimulus. The spike trains were used as inputs to the reservoir and different states were recorded for post-processing. One state corresponds to the membrane potentials of all the neurons in the reservoir. In this experiment, where a reservoir of size 8 neurons is used, total eight membrane potentials were recorded in one state. Total 20 spike trains were generated for one digit and 200 spike trains were processed through the reservoir for total 10 digits. The 'readout' neurons (feedforward network) were implemented in software for the classification of input digits. Fig.5 shows an input digit1 and corresponding spike trains.

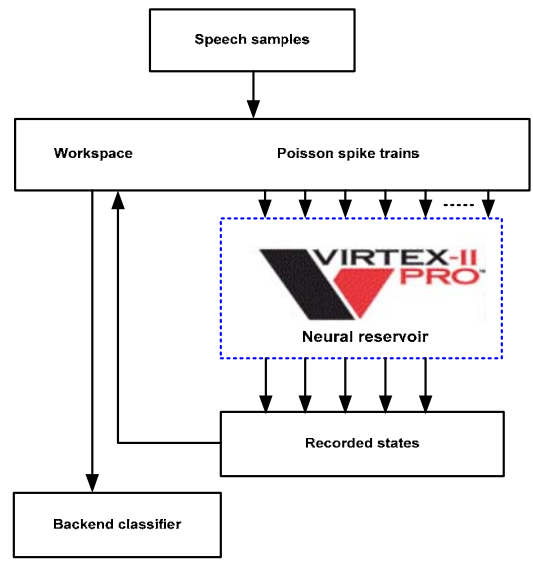

**Fig. 4.** An integrated environment for neural reservoir based recognition

The network is constructed in a way that each neuron was connected with minimum two inputs where input spikes were weighted through fixed weights. The fixed weights were compared with the numbers generated by the

LFSR. Once the required synaptic efficacy is achieved for the incoming spike, an output value of '1' is generated which is further reduced with three bits shift right operation. This is important because neuron threshold voltage was set to 0.15 V. The complete architecture consists of an LFSR to generate random numbers, delayed registers to feedback some of the outputs back to the inputs and LIF neurons which consists of synapse and membrane units.

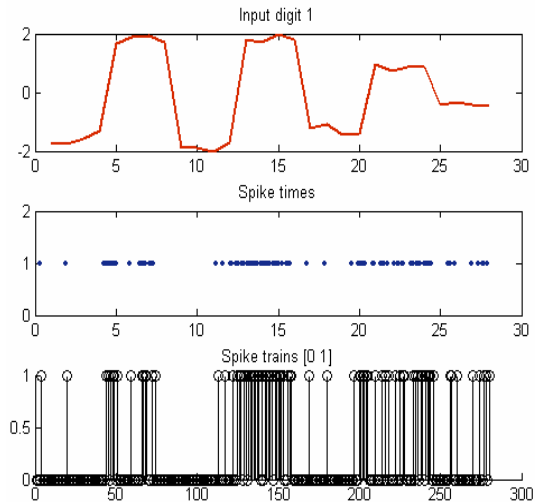

**Fig. 5.** Input digit and corresponding spike times and spike trains

In order to interface the input spike trains with neural reservoir, they have to be converted into Simulink Boolean type through input gateways. These input spike trains were used to perturb the reservoir and the responses were collected in terms of membrane potentials and stored in Matlab workspace for backend classification. The reservoir has to be simulated for the total time steps equivalent to the time steps of spike trains in order to feed input data into the reservoir. Once the total states were recorded, they were further sampled and only five states were recorded for post-processing, the states were recorded in linear scale from start to the end of states. The readout neurons were trained offline until the algorithm converged to the goal and tested to evaluate their classification accuracy.

A three-layered recurrent neural reservoir (3x2x3) is implemented (see fig.6) where input vectors were directly connected to the neurons and each neuron had minimum two synapses. Total 8 neurons with 16 synapses were implemented. It is possible to increase the number of neurons and synapses with a chain of adders for synaptic accumulation. All the neurons in the reservoir work in parallel because all inputs and corresponding random weights were accessed simultaneously. The size of the network can be increased with the increased number of neurons and synapses. The input spike trains were connected with the neurons in the network because connectivity is an important factor for reservoir to have 'short term' memory. Total 16 fixed weights were stored for 8 neurons in the network where each neuron had minimum two inputs. The input stimulus to the reconfigurable reservoir was the Poisson spike trains which were generated off-chip. The reservoir states (membrane potentials) were also stored off chip for backend classification. Once the reservoir is simulated, the total states of the reservoir were further sampled and five states were recorded for post-processing. The states were sampled in linear scale from start to the end in linear scale (50,100,150, 200 and 250). These five frames were further processed for backend classification.

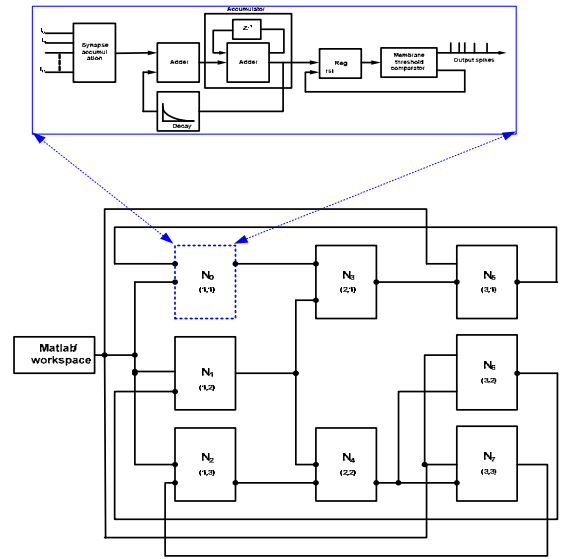

**Fig. 6**. Reconfigurable neural reservoir

An MLP classifying engine was implemented as backend in software. Total data was split into two sets: training and testing. One training sample consisted of 40 data points (5 states of total number of eight neurons at five linear time steps from start to the end) and 10 output neurons were used in the output layer which corresponds to the 10 isolated digits. The network was trained with the training samples where each training sample was compared with the target and the accuracy is calculated. The total number of 30 hidden neurons were used which was found to be the best combination with input layer neurons. The overall accuracy drops if the number of hidden layer neurons were increased or decreased to the maximum number of 30 neurons. After testing with the test data and different hidden layers an overall accuracy of 98% was achieved on test data and 100% accuracy was achieved on training data set. Different standard backpropagation training algorithms were tested but best results were achieved when the network was trained with Matlab Levenberg-Marquardt training algorithm. The training took 124 seconds and converged to the goal after 25 iterations (see fig.7).

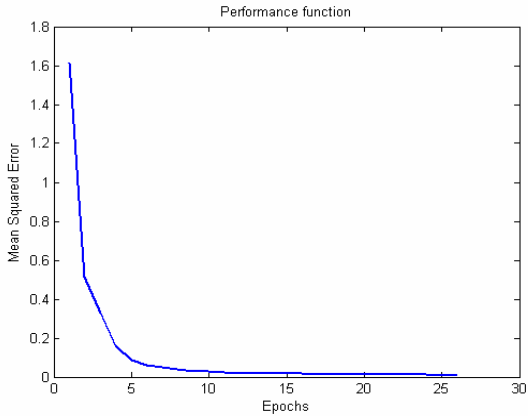

**Fig. 7**. Total number of epochs and mean squared error

## 5. CONCLUSION

This work has proposed a novel HW/SW environment for developing a reservoir based approach on reconfigurable platforms. This paper has contributed in presenting a design strategy for implementing neuro inspired systems on reconfigurable platforms. A novel reconfigurable hardware architecture is presented and resources were calculated. It is shown that a multiplier-less synapse processing is possible and the multipliers required for synapse modelling can be avoided completely. The architecture is scalable and can easily be scaled on multiple FPGAs to form distributed compact parallel reservoirs. The proposed strategy has demonstrated that it is possible to map compact neural reservoir with low area requirements and demonstrated with speech recognition application.

For the hardware implementation, fixed point simulations were used and in order to save the overall area, the range of fixed points was kept to the minimum. The proposed parallel architecture describes a node parallelism where each node has its own functional unit for neuron processing. The results of the hardware/software based implementation differs less than 2% of the floating point pure software implementations.

The proposed architecture and overall system provides several distinct advantages because circuit size is significantly optimised and secondly no control circuitry is required. Each cell is implemented as standalone computing unit and interconnected with neighboring neurons. These advantages make it possible to design and implement large self contained neural reservoirs on reconfigurable platforms. The results obtained from the proposed hardware/software implementation of neuro inspired paradigm are promising and opens new possibilities to build intelligent systems.